# Depth image hand tracking from an overhead perspective using partially labeled, unbalanced data: Development and real-world testing

Stephen Czarnuch and Alex Mihailidis

*Abstract*—We present the development and evaluation of a hand tracking algorithm based on single depth images captured from an overhead perspective for use in the COACH prompting system. We train a random decision forest body part classifier using approximately 5,000 manually labeled, unbalanced, partially labeled training images. The classifier represents a random subset of pixels in each depth image with a learned probability density function across all trained body parts. A local mode-find approach is used to search for clusters present in the underlying feature space sampled by the classified pixels. In each frame, body part positions are chosen as the mode with the highest confidence. User hand positions are translated into hand washing task actions based on proximity to environmental objects. We validate the performance of the classifier and task action proposals on a large set of approximately 24,000 manually labeled images.

## I. Introduction

IN this paper we describe the development and testing of a new hand tracker for the COACH automated task prompting system [1, 2] using depth images captured using a Kinect [3] sensor mounted above a washroom sink. The COACH [1, 4] is an automated prompting system designed to facilitate independent completion of daily activities (e.g., hand washing) by older adults with dementia. The system employs an automated planning algorithm, called a partially observable Markov decision process (POMDP) [5] to infer task progression. The hand washing task is divided into five fundamental steps or behaviors [1], where some steps must be completed in an ordered sequence while others can be completed in any order. The POMDP translates the primitive actions of users into a measure of their progression through the task, allowing for ordered and unordered step completion, while accounting for the partial observability, or unreliability, of the sensor readings identifying the users' actions.

For the task of hand washing, data input to the POMDP planning module is the user's hand locations relative to environmental objects. Hand locations can take on one of six positions: in the sink, adjusting the faucets, rinsing in the water, using the soap, using the towel, or away from the sink [2]. To determine these positions, the COACH employs an overhead computer-vision tracking algorithm that identifies the hand positions in two dimensions [6]. Prior to using the system each of the six environmental positions is statically defined. The hand positions are then translated to one of the six environmental positions based on their proximity to the predefined environmental regions in the image.

The COACH hand-tracking system has progressed through three technical generations of development [2, 4, 7-16] to its current state that employs a color-based flocking algorithm [15]. A controlled, clinical evaluation of the COACH system, utilizing the color flocking algorithm, suggested that color-based hand tracking was effective and that the system could follow the actions of its users through the hand washing task [4]. These results motivated a needs assessment study to understand the role of the COACH in the lives of older adults with dementia and their caregivers, toward an ultimate real-world deployment [17]. To ensure the COACH was satisfying these needs, the system was deployed in an unsupervised state in a dementia treatment facility in Toronto, Canada[1]. This deployment revealed that color-based hand tracking could not accurately track the actions of a user if the user was not known a priori. Additionally, the study showed that color-based tracking could not easily separate different body parts that were similar in color. Examples of this include times when a person was wearing skin-colored or sleeveless clothing, or was partially or fully bald. These conditions are particularly challenging when considering that the images are captured from an overhead perspective, where often only the head, shoulders, arms and hands of the users are visible. Most notable, however, was that the color-based hand tracking could not disambiguate the hands passing over objects in the environment (e.g., taps, soap dispenser) versus the hands actually interacting with the objects. Unreliable hand tracking significantly affected the overall performance of the COACH system as a whole [1]. However, the trials were simulated using manually constructed ground-truth hand tracking data. The simulations resulted in excellent overall system performance [1], suggesting that the color-based hand tracking was restricting the overall performance of the COACH.

Stephen Czarnuch is with the Institute of Biomaterials and Biomedical Engineering, University of Toronto, Toronto, ON Canada. (e-mail: stephen.czarnuch@utoronto.ca).

Alex Mihailidis is with the Department of Occupational Science and Occupational Therapy and Institute of Biomaterials and Biomedical Engineering, University of Toronto, Toronto, ON Canada, (e-mail: alex.mihailidis@utoronto.ca).

To ensure that improvements to the hand tracking would ultimately fulfill the needs of the users of COACH a House of Quality (HOQ) approach [18, 19] was employed. HOQ is a systematic methodology used to convert customer needs into design specifications. Using an HOQ approach, the needs of the users of COACH [1] were weighed against the technical capabilities of the COACH in a real world deployment [17]. The results of the HOQ analysis motivated the development of a more accurate and reliable tracking approach that would improve the overall performance of the COACH system.

*A. Related Work*

The challenge of obtaining reliable, unobtrusive human body tracking data is not unique to the COACH system, affecting diverse areas such as gaming, human-computer interaction, and health care [20]. The development of the Kinect [3] sensor and software, largely in response to increased gaming requirements, provided an approach to human posture recognition and tracking that overcame a substantial number of limitations experienced by previous tracking systems [20]. The approach utilized depth imaging to accurately provide uninitialized (frame by frame), three-dimensional body part proposals that were largely color, texture, shape and lighting invariant [21]. The methodology forms a core component of the Microsoft Kinect [3] gaming platform which has been successfully deployed in the homes of many users around the world.

The methodology of Shotton et al. [20] has been used directly to classify static, full-frame hand and finger poses [22]. Furthermore, proposed improvements to various elements of the approach, geared toward increased classification accuracy and speed, have been discussed [23, 24]. Kohli and Shotton [25] have even extended the earlier work of Shotton et al. [20] to remediate problems associated with the approach.

Depth-based tracking development has not been restricted to extensions of the methodology of [20]. Tracking using range cameras (e.g., structured light like the Kinect [3] and time of flight [see 26 for details]) have been used to successfully identify human poses and positions [27, 28], and distinguish hand poses from a fixed frontal [22, 29] and egocentric [30] perspective. Depth-based tracking has also been used from a fixed frontal perspective to identify hand and head positions [31], body positions [32], upper body segmentation [24], and full joint and body part positions [21, 23]. The task of body part classification from a frontal perspective became significantly easier through the release of the Kinect SDK and the development of open source APIs [e.g., 33].

*B. Existing Tracking Approaches and the COACH*

Three significant limitations prevent the application of existing depth-based hand and body tracking approaches to the COACH. The first limitation is that the COACH system utilizes an overhead, birds-eye tracking perspective to ensure an unobstructed view of its users and unobtrusive installation. The COACH system is targeted at older adults with dementia who are cognitively compromised requiring that any installed hardware remain out of reach. Furthermore, caregivers of older adults with dementia have indicated that any assistive technologies must integrate into the environment to reduce the likelihood of stigmatization [17]. To our knowledge, existing approaches have not attempted part tracking and recognition from a fixed overhead perspective using depth imaging. Rather, most approaches utilize a frontal perspective [20, 23, 24, 27-29, 32], and/or perspectives unique to a particular application [22, 24, 29]. The second limitation is that many existing tracking approaches require the entire object to be in the scene in order to initialize the tracking model. In the case of the COACH system, users are often only partially in the scene when the hand washing task begins, specifically when washrooms are small. The final limitation preventing the use of existing tracking approaches is that existing approaches implement tracking from either a global or a local perspective. Global perspectives require fully labeled training data over all body parts and typically require substantial training sets to provide reliable part tracking over the entire range of possible human motions [e.g., 20]. The generation of large, fully labeled data sets is time consuming and prone to labeling error. Local perspectives, on the other hand, focus on specific challenges such as hand or finger poses, or the reduction of all possible body positions to a small number of predefined poses [e.g., 27, 28]. The task of hand washing, moreover, is inherently variable, preventing the direct use of poses or posture as an indication of task progression. However despite existing limitations, the work proposed by Shotton et al. [20] presented a general framework for training a body part and joint classifier using single depth imaging.

*C. Contributions*

Given that existing approaches have not attempted part tracking from a fixed overhead perspective, our main contribution is the development of an overhead hand tracking methodology using single depth imaging with application to the COACH prompting system. In the process, we have developed a method of training an uninitialized, frame-by-frame classifier, based on the work of Shotton et al. [20], using unbalanced, partially labeled training data captured from an overhead perspective. These unbalanced training data consist of a small number of labeled image pixels relative to the total number of pixels in the training image. We then show that the classification accuracy of the part tracking is sufficient to integrate directly into the COACH system as a hand tracker without a temporal or kinematic model. Finally, we validate the COACH's ability to track the task-based activities of the participants through the hand washing task using the depth-based hand tracker on a large set of supervised, real-world trials.

Through this work we seek to answer the following three research questions related to the new depth-based overhead hand tracker trained using unbalanced, partially labeled training data:

1. What effect will partially labeled training data have on the training and performance of decision trees?

2. What training parameters impact the overall performance of the classifier?
3. What is the mean average precision of an optimal classifier when proposing hand positions in three dimensions compared to ground-truth hand positions?
4. How accurately can the COACH track the completion of task-based activities of users through the task of hand washing in a supervised real-world environment?

## II. METHOD

The research of Shotton et al. [20] provided the basis for the approach we use to classify individual body parts from a single overhead depth image on a per-frame basis. The acquisition and development of our unique training and validation data are captured from a fixed overhead perspective. We first generate a random decision forest using a simple depth feature to provide intermediate multiclass probability density functions (PDF) for each sampled image pixel. We then propose final body part positions by aggregating the information contained in the underlying PDF. We evaluate the performance of the intermediate and aggregate classifiers and optimize key training parameters. The optimal parameters are used to train a final decision forest resulting in a new depth-based hand tracker. We evaluate the final hand tracker against ground-truth hand positions and integrate the tracker into the COACH system. Finally, we evaluate the COACH's ability to track the task-based activities of the participants using a set of validation images.

### A. Data Collection

Study participants were recruited from a pool of researchers associated with Toronto Rehabilitation Institute [34] on a voluntary basis. An overhead Kinect sensor recorded depth and RGB images of participants washing their hands in a fully functional washroom located in Toronto Rehabilitation Institute's HomeLab [35]. Both depth and RGB images were captured at 30 frames per second and a resolution of 640 x 480. A random subset of the recorded trials was removed from the data set and used as training data for the classifier. A second random subset, independent of the first, was also removed from the data set and used as a holdout image set to measure the performance of the classifier. The remaining data were used to validate the performance of the COACH system.

The foreground from each image in the training data set was removed from the background using simple depth thresholding, while maintaining any pixels identified as invalid by the sensor in both the background and depth image. Each foreground image pixel was assigned to a class $p$ = {*left hand; right hand; head; body*}. Only the *left hand, right hand* and *head* were directly labeled; any remaining pixels in the foreground were assumed to be *body*. The labeled images composed the set of images used to train the classifier. The same foreground extraction and labeling methodology was employed to create the independent set of holdout images used to evaluate the classifier performance. The ground-truth centers for each part in each holdout image were calculated as the center of mass of each body part.

For each validation image, the participant's action was assigned one of a mutually exclusive set of *action* = {*walking; washing hands; drying hands; turning*}. Walking was defined as any frame where the participant was approaching or leaving the washing area along the normal path from the door of the washroom to the sink area. Washing hands was defined as any frame where the participant was facing the sink area with any body part over the counter. Drying hands was defined as any frame in which the participant's hand was in contact with the towel and not turning. Turning was defined as any frame where the participant was not facing the sink area, and not *walking*. Furthermore, for each image, the participant's task activity was identified as one of *activity* = {*away; soap; tap; water; sink; towel*} according to our previous work [2]. Finally, for each image the center of the each part was labeled. An annotation tool was developed to label the part centers in the validation images. To reduce annotation time, the center of each part was initially estimated by the classifier and manually corrected if required.

### B. Intermediate Multiclass Classification

An intermediate multiclass classification was implemented, based on the work of Shotton et al. [20], which represented each of a random subset of pixels across a depth image $I$ as a learned probability density function across all body parts. The background segmented and labeled training images $I_t$ were used to train a set of random decision trees. A set of training samples S = {$I_t$, **x**} were determined by randomly proposing a set of $N$ pixels **x** from each training image. Then, a set of offset vectors $\theta$ = (**u**, **v**) and thresholds $\tau$ were randomly generated with maximum magnitude $\theta_{max}$ and $\tau_{max}$ respectively. Together, the offset vectors and thresholds created a set of splitting criteria $\phi$ = ($\theta$, $\tau$). A simple depth feature from [20] was then defined as

$$f_\theta(I_t, x) = d_{I_t}\left(x + \frac{u}{d_{I_t}(x)}\right) - d_{I_t}\left(x + \frac{v}{d_{I_t}(x)}\right) \quad (1)$$

where $d_I(x)$ was the depth at pixel **x** in image $I$. The resulting depth feature was then calculated as the difference between depth probes at the points identified by each offset vector **u** and **v**. To ensure that the offset vectors were invariant to depth, vectors **u** and **v** were normalized by the depth of pixel **x** providing the same world-space offset regardless of the distance of the pixel from the sensor. In the case where the pixel identified by an offset probe was located outside on the background or off the image, the depth probe returned a large positive value greater than the maximum depth possible from a valid pixel.

A set of binary decision trees was trained using training sets unique to each tree. Training started at the root node with all training samples, and each node $i$ of the tree was determined according to the following steps:

1. Divide the set of training samples $S_i$ at node $i$ into left and right subsets for each split candidate/threshold pair $\phi$ according to the depth feature, where

$$S_{i,L}(\phi) = \{I_t, \mathbf{x}\} \mid f_\theta(I_t, \mathbf{x}) < \tau \quad (2)$$

$$S_{i,R}(\phi) = S_i \setminus S_{i,L}(\phi) \tag{3}$$

2. Determine the $\phi$ which removes the purest left and right subsets, providing the largest information gain

$$g(\phi) = H(S_i) - \frac{|S_{iL}|}{|S_i|} S_{i,L}(\phi) - \frac{|S_{iR}|}{|S_i|} S_{i,R}(\phi) \tag{4}$$

$$\phi_{max} = \arg\max(g(\phi)) \tag{5}$$

where $H(S_i)$ is the entropy of the normalized PDF of $S_i$.

3. If $g(\phi_{max})$ is greater than a minimum gain $g_{min}$ and the tree depth has not exceeded the maximum depth $D_{max}$, recursively repeat steps 1 to 3 for the left $S_{i,L}(\phi_{max})$ and right $S_{i,R}(\phi_{max})$ subsets.
4. If $g(\phi_{max})$ is less than a minimum gain $g_{min}$ or the tree depth has exceeded the maximum depth $D_{max}$, store the probability density function $P_i(p|I_t,\mathbf{x})$ across all parts.

Decision trees tend to overfit data, particularly when training sets are small [20]. Accordingly, the decision trees were used in an ensemble of $T$ trees to form a decision forest. For a given pixel $\mathbf{x}$, each tree was traversed until a leaf node was reach. The final PDF was given as the average of the individual PDFs provided by each tree

$$P(p|I,\mathbf{x}) = \frac{1}{T}\sum_{n=1}^{T} P_n(p|I,\mathbf{x}) \tag{6}$$

We measured the impact of five training parameters on the multiclass classification, or per-pixel, performance of the intermediate classification: 1) the number of training images; 2) the maximum tree depth $D_{max}$; 3) minimum cutoff gain $g_{min}$; 4) maximum magnitude of the offset vectors $\theta_{max}$; and 5) number of samples per image $N$. A forest was trained by varying each parameter while holding all other parameters constant. A confusion matrix was generated for each forest between the most likely predicted body part and the ground-truth part using a set of labeled holdback images. Similar work by [20] and recommendations for multiclass classifier performance [36] measurement suggest the use of the Overall Success Rate (OSR), or simply the sum of the diagonal of the confusion matrix. This measure works well for balanced data sets, but partially labeled data sets are proportionally imbalanced. Accordingly, the intermediate part classification performance was calculated as the unweighted average recall (UAR) [37] of each class in the confusion matrix to account for the unbalanced classes.

*C. Aggregate Part Proposals*

For each image in the holdback set ($I_h$), the random decision forest provided an intermediate classification of the set of $N$ pixels, where each pixel was assigned a PDF across all body parts. The PDF of each sampled pixel was taken as a representative point of an underlying density function or feature space. Final body part position proposals provided the most likely position of each part by searching for clusters present in the feature space. The coordinates of each pixel $\mathbf{x}$ were transformed from projective space into pixel $\acute{\mathbf{x}}$ in world space creating a 3D density function for each body part. Initial estimates for each part $\hat{x}_p^0$ were identified as any pixel with a part probability above a threshold $\varphi_p$

$$\hat{x}_p^0 = \{I, \acute{\mathbf{x}}\} \mid P(p|I,\mathbf{x}) > \varphi_p \tag{7}$$

The maxima $\hat{x}_p$ of each density function were then determined for each part using a mean shift mode seeking algorithm [38] with a weighted Gaussian kernel $w_p$ from [20]

$$w_p = P(p|I,\acute{\mathbf{x}}) \cdot d_I(\acute{\mathbf{x}})^2 \tag{8}$$

$$\hat{x}_p^{t+1} = \frac{\sum_{i=1}^{N} w_p \cdot \acute{\mathbf{x}}_i \cdot e^{\frac{|\hat{x}_p^t - \acute{\mathbf{x}}_i|^2}{h_p^2}}}{\sum_{i=1}^{N} e^{\frac{|\hat{x}_p^t - \acute{\mathbf{x}}_i|^2}{h_p^2}}} \tag{9}$$

where $h_p$ is a per-part bandwidth that influences the convergence rate and final number of part proposals. The confidence of each mode was give as the sum of the contributions of each pixel to the final mode

$$\alpha_p = \sum_{i=1}^{N} w_p \cdot e^{\frac{|\hat{x}_p - \acute{\mathbf{x}}_i|^2}{h_p^2}} \tag{10}$$

The mode with the highest confidence was used as the final part position proposal.

We evaluated the aggregate part proposal performance of the mode-finding algorithm over the set of holdback images given a starting threshold ($\varphi_p$). The distance between all proposed modes for each part was compared in world-space to the ground-truth part centers. For each part, a confusion matrix was constructed between the ground-truth part position and the proposed position. A true positive (TP) was scored for the first mode within a minimum per-part distance $\Delta_p$ of the ground-truth part center. Any other mode within $\Delta_p$ or any mode outside $\Delta_p$ was scored as false positive (FP). A true negative (TN) was scored if the part did not exist in the image and the classifier did not propose any modes. A false negative (FN) was scored for all modes proposed in an image that did not contain a corresponding part.

Precision-recall (PR) curves were generated for each body part as a function of mode find starting threshold ($\varphi_p$) over the holdback image set. The average precision (AP) was calculated for each part. The aggregate part proposal performance was calculated as the mean average precision (mAP) across all parts. The optimal starting threshold ($\hat{\varphi}_p$) was determined for each part using the Equal Error Rate – the point where precision and recall were equal.

*D. Task-based Hand Positions*

Modes were provided by the classifier for each part for each image in the validation image set using the final decision forest and starting thresholds ($\hat{\varphi}_p$). The mode with the highest confidence ($\alpha_p$) was u as the final part proposal. A confusion matrix was constructed between the ground-truth part position and the proposed position in world space for each part. For each image, a true positive (TP) was scored if the proposal was within a minimum per-part distance $\Delta_p$ of the ground-truth part center. If the proposal was outside $\Delta_p$ it was scored as false positive (FP). A true negative (TN) was scored if the part did not exist in the image and the classifier did not propose any modes. A false negative (FN) was scored if a position was proposed for the part in an image that did not contain a corresponding ground-truth part. An $F_\beta$ measure was used to evaluate the classifier performance, where $\beta$ is a factor weighting the importance of precision and recall

$$F_\beta = \frac{(1+\beta^2) \cdot precision \cdot recall}{\beta^2 \cdot precision + recall} \tag{11}$$

An $F_{0.5}$ measure was calculated for each part $p$ in each validation each category *action* the validation data set. The scaling factor of $\beta = 0.5$ was selected to prioritize the correctness of any classified parts. The final $F_{0.5}$ measure per part was calculated as the mean measure over all trials.

For each frame in each validation trial, the left and right hand positions (determined by the mode with the highest classification confidence) were converted to one of the six participant task activities defined by *activity*. Each *activity* was characterized by a pre-defined spheroid in world space. When a hand entered a region defined by a spheroid and persisted for three or more consecutive frames the associated *activity* was considered active for use by the POMDP decision policy.

A confusion matrix was created for each validation trial to measure the COACH's ability to track each hand washing *step* = {*turn on water, get soap, rinse hands, turn off water, dry hands*} [4]. A *step* was considered complete if the associated *activity* occurred in sequence. A true positive (TP) was scored when a user completed a task activity and the COACH correctly identified the activity as complete. A true negative (TN) was scored if a task activity was not completed and the COACH did not indicate the activity was completed. A false positive (FP) was scored if a task activity was not completed but the COACH identified the activity as completed. A false negative (FN) was scored if a task activity was completed but the COACH did not indicate it as correctly completed. An $F_1$ measure from (11) was calculated for each trial, equally weighting precision and recall. The overall performance of the system was measured as the average $F_1$ measure over all trials.

## III. RESULTS

### A. Data Collection

Depth images were captured over fifty-one (51) trials where thirty (30) unique participants performed a real hand washing task. Trials were conducted in a fully functional washroom located in Toronto Rehabilitation Institute's HomeLab [35]. Participants were associated with the research team and were included in the study on a voluntary basis. A total of 34,999 frames test data were collected. After removing all frames that did not contain a foreground image, a total of 28,763 frames of useable data remained. The training data set ($I_t$) was comprised of eight (8) trials and a total of 4,971 frames of depth data. The holdout data set ($I_h$) included two (2) trials and a total of 854 images. The validation data set ($I_v$) included the remaining forty-one (41) trials and a total of 22,938 images.

### B. Intermediate Multiclass Classification

Initial decision trees were trained using the following default parameters: 100 thresholds ($\tau$), 3000 offset vectors ($\theta$), maximum offset ($\theta_{max}$) 500, minimum gain ($g_{min}$) = 0.05, 4000 samples per image ($N$), maximum tree depth ($D_{max}$) = 20 and 4971 training images ($I_t$). Final decision tree parameters were identified by varying a single parameter and setting all others to the default value. Figure 1 shows the results of the parameter optimization training. Investigating each chart suggests that the forest achieves a maximum accuracy at $D_{max}$ = 12, $I_t$ = 4971, $g_{min}$ = 0, $\theta_{max}$ = 250 and $N$ = 3000.

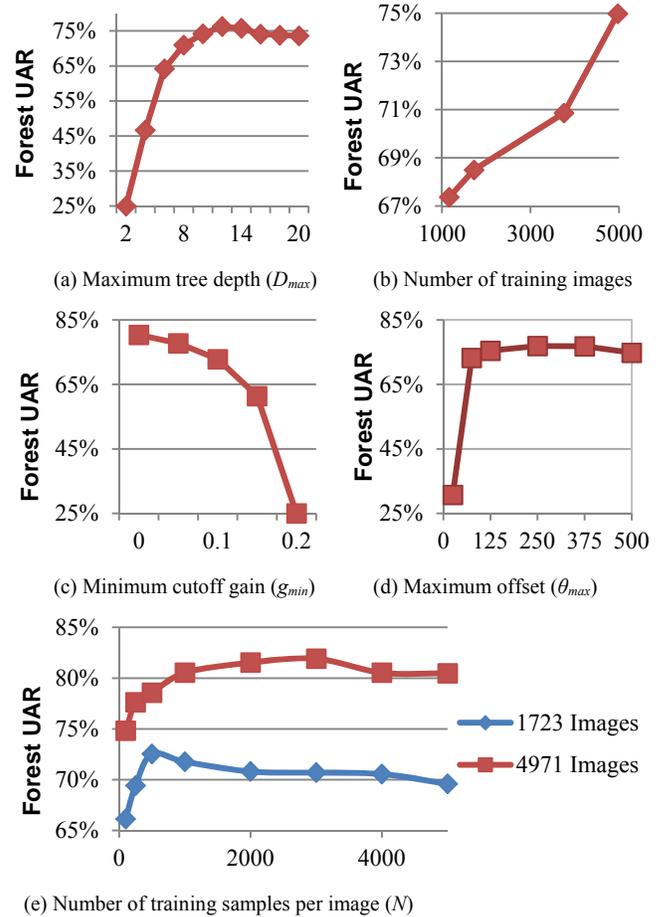

(a) Maximum tree depth ($D_{max}$)
(b) Number of training images
(c) Minimum cutoff gain ($g_{min}$)
(d) Maximum offset ($\theta_{max}$)
(e) Number of training samples per image ($N$)

Figure 1: Unweighted average recall (UAR) versus training parameter. (a) Depth of decision tree. (b) Number of training images per forest. (c) Minimum cutoff gain for leaf node. (d) Offset vector maximum offset (in pixel-meters). (e) Number of training sample pixels per image: 1723 and 4971 training images per forest.

### C. Aggregate Part Proposals

The holdout image set ($I_h$) was used to generate precision-recall (PR) curves for each part over the range of starting threshold probabilities $0 \leq \varphi_p \leq 1$. Table 1 shows the average precision (AP) and optimal starting threshold ($\hat{\varphi}_p$) for each part, and a mean average precision mAP = 0.846. Notable is the high starting threshold value $\hat{\varphi}_p = 0.95$, suggesting that the intermediate classification of the "head" is reliable.

**Table 1:** Aggregate part proposal performance parameters and outcomes

|  | *Left Hand* | *Right Hand* | *Head* |
|---|---|---|---|
| **Minimum distance ($\Delta_p$)** | 5 | 5 | 10 |
| **Average precision (AP)** | 0.802 | 0.805 | 0.931 |
| **EER threshold ($\hat{\varphi}_p$)** | 0.65 | 0.6 | 0.95 |

*D. Task-based Hand Positions*

Table 2 shows the mean $F_{0.5}$ measure for each body part *p* within each *action* category, along with the number of frames within each *action* category from the set of validation and training data sets. The mean $F_{0.5}$ measure was calculated as the average $F_{0.5}$ score over all validation trials for each body part using the EER starting thresholds ($\hat{\varphi}_p$). Over all validation frames, the $F_{0.5}$ measure for the left hand, right hand, and head were 0.713, 0.789 and 0.890 respectively. In all cases classification of the head outperforms classification of the left and right hands except in the *walk* action where the head is often not in the scene. Additionally, classification of all body parts substantially underperforms during the "walk" and "turn" *actions* compared to the "wash" and "towel" *actions*.

A disproportionately large number of training images were in the "wash" *action*, represented by 697 (81.6%) of all training images. Conversely, only 32 (3.8%) and 29 (3.4%) of all training images were in the "turn" and "towel" *action* categories respectively. In comparison, a total of 15,613 (68.1%) validation images were in the "wash" *action*, while 1,505 (6.6%) and 2,968 (12.9%) were in the "turn" and "towel" *actions* respectively.

**Table 2:** Mean final part proposal $F_{0.5}$ measure of performance over the set of validation trials, categorized by *action* and body part *p*; number of validation and training frames within each *action* category.

| Action | Body Part (*p*) | Mean $F_{0.5}$ Measure | Validation Frames (%) | Training Frames (%) |
|---|---|---|---|---|
| **Walk** | Left Hand | 0.219 | 2852 (12.43%) | 96 (11.24%) |
|  | Right Hand | 0.315 |  |  |
|  | Head | 0.305 |  |  |
| **Wash** | Left Hand | 0.879 | 15613 (68.07%) | 697 (81.62%) |
|  | Right Hand | 0.908 |  |  |
|  | Head | 0.997 |  |  |
| **Turn** | Left Hand | 0.082 | 1505 (6.56%) | 32 (3.75%) |
|  | Right Hand | 0.199 |  |  |
|  | Head | 0.497 |  |  |
| **Towel** | Left Hand | 0.355 | 2968 (12.94%) | 29 (3.4%) |
|  | Right Hand | 0.699 |  |  |
|  | Head | 0.994 |  |  |
| **All** | Left Hand | 0.713 | 22938 (100%) | 854 (100%) |
|  | Right Hand | 0.789 |  |  |
|  | Head | 0.890 |  |  |

The performance of the COACH as an automated task support over the validation image set is summarized in Table 3. Based on the data presented in Table 3, the precision of the system is 0.994 and the recall is 0.938. The resulting $F_1$ measure score is 0.965.

**Table 3:** The performance of the COACH while monitoring the task-based activities of the participants over the set of validation trials.

|  |  | COACH | |
|---|---|---|---|
|  |  | Activity Identified | Activity Not Identified |
| **Participant** | Activity Completed | 180 | 12 |
|  | Activity Not Completed | 1 | 12 |

IV. DISCUSSION AND CONCLUSION

The purpose of this study was to develop and integrate a novel hand tracker into the COACH prompting system using depth images captured from an overhead perspective. Existing depth-based body part, pose and posture tracking methodologies were not suitable for this application. The work of Shotton et al [20] was adaptable to our application, however using the approach required a substantial amount of training data in order to develop a functional classifier.

We captured a large set of depth and RGB images, totaling 28,763 usable frames of video. However, manually creating fully labeled training and holdback data sets was time consuming, taking three to five days for each trial. To reduce manual labeling time, we employed a partial labeling approach where we only labeled body parts with application to the COACH prompting system. To further reduce the data preparation time, we simply labeled the part centers for the validation image set.

Using partially labeled data to train decision trees in a random decision forest provided comparable results to the foundational studies we used to develop our methodology [e.g., 20, 23, 24]. The intermediate multiclass classification accuracies we achieved were comparable to the results presented in those studies. This is likely due to the fact that the Shannon entropy used to determine the optimal split criteria is robust to proportionally unbalanced probability density functions. These results suggest that fully labeled training data are not necessary when training a random decision forest. Furthermore, when evaluating the performance of a decision forest, the unweighted average recall [37] is an appropriate and effective measure for an unbalanced data set.

Varying the values of training parameters and evaluating the performance of the resulting intermediate classifier clearly showed the optimal values for those parameters or the trend caused by varying the parameters. In the case of the maximum tree depth ($D_{max}$), a reduction in classification accuracy was visible in Figure 1(a) with depths beyond the optimal value. This is likely a reflection of the tendency decision trees have to overfit to the data. A similar trend is visible when considering the maximum offset ($\theta_{max}$) in Figure 1(d). Shotton et al. [20] propose that this is likely due to overfitting to an increase spatial context. We also propose that increasing the maximum magnitude of the offset vectors increases the likelihood that an offset vector will generate an offset probe outside the bounds of the image. All offset probes landing outside the image result in the same probe value. Increasing the number of vectors consistently reaching outside the image functionally reduces the total number of unique offset vectors usable by the training routine. Increasing the number of training samples ($N$), shown in Figure 1(e), beyond the optimal value also leads to a decrease in the classification accuracy of the system. This is largely due to overfitting the decision trees to the training data. A clear trend is visible in Figure 1(b), indicating that an increase in the number of training images increases the classifier accuracy, which is in agreement with similar studies [20]. Finally, setting a minimum cutoff gain ($g_{min}$) exponentially reduces the accuracy of the classifier (Figure 1(c)). The advantage of increasing the minimum cutoff gain is that the memory penalty is reduced. Larger gains result in less complex decision trees. In the case of our forest the size of each decision tree is reduced by a factor of approximately $2^n$, where $g(1 … n) = (0, 0.05, 0.1 … n)$. However, the overall size of our trees, trained with four body parts, is relatively small. If the complexity of the trees is increased by classifying more parts or by adding additional training images, sacrificing accuracy to reduce the memory penalty may be worthy of consideration.

Aggregating the underlying probability density functions provided by the intermediate classifier effectively provided final part position proposals. In particular, proposals for the position of the *head* were given with an average precision of 0.931. Interestingly, the optimal threshold to identify starting points for the mode find algorithm was 0.95 for the "head", suggesting a high confidence in the intermediate classification for that part. In contrast, the starting thresholds for the "left hand" and "right hand" were 0.65 and 0.60 respectively, indicating a lower confidence in the intermediate classification. Regardless, the aggregate left and right hand position proposals still resulted in an average precision of 0.802 and 0.805 respectively. This is not surprising as the "head" is much less variable in terms of position, shape and size than the "left hand" and "right hand".

The performance of the final hand tracker over the set of validation trials yielded highly variable results (Figure 2). Overall, the system performed well based on high $F_{0.5}$ measures for each body part. However, evaluating the $F_{0.5}$ measures for each *action* of the participants shows that the classifier didn't perform as well in some cases. In particular, the *action* "walk" and "turn" performed poorly for all body parts. Additionally, classification of the "left hand" was not overly accurate during the "towel" *action*. This is explainable by investigating the proportion training and validation image frames in each *action*. A disproportionately large number of training image frames (81.62%) were in the "wash" *action*. Additionally, only 11.24% of all training image frames were in the "walk" *action*, with the remaining images split evenly between the "turn" and "towel" *action*. This disproportionate distribution among *action* categories resulted in a decision forest biased toward classifying body parts in the "wash" *action*. Indeed, the relatively decent performance of the classifier during the "towel" *action* is likely a coincidental result of the similarity between human motions during the "wash" and "towel" actions. The other "walk" and "turn" actions, in comparison are much more unique and as a result provided poorer classification performance.

The COACH system performed exceptionally well at correctly identifying the participants' activities over the 22,938 validation images. Over the forty-one (41) trials, participants could have potentially completed 205 task steps. Of the 192 completed steps the system correctly identified 180 as complete. Of the remaining thirteen (13) steps that were not completed, the system correctly identified twelve (12) as not completed. Considering the thirteen (13) incorrectly classified activities: seven (7) occurred because the classifier failed to correctly identify the locations of the hands for the duration of the activity; three (3) were the result of the hands completing the action outside the defined region; two (2) occurred when the hands were fully obscured from the camera's view for the duration of the activity (e.g., user leaning over the sink); and the last was a system issue where several frames were not recorded properly preventing proper classification of the hands. A critical point to address when considering the success of the hand tracker is the paucity of False Positive results. Depth tracking was easily able to disambiguate the completion of certain task actions compared to the hands simply passing over action *activity* regions. False Positives triggered by the hands moving above *activity* regions was a significant contributor to poor performance in our previous colour-based tracker [1].

Despite measurable gains in intermediate classification through individual parameter optimization, our study did not investigate the potentially complex interactions between all the training parameters. Future work will look to further improve intermediate classification by concurrently optimizing training parameters. Furthermore, the most significant gains are likely achievable by increasing the training data set – particularly by including more images in task *actions* that underperformed during the validation trials. The value of a larger training set is perhaps most supported when considering the results achieved by Shotton et al. [20], who used an enormous training set of 900,000 real and synthetic images. From a more practical perspective, in some instances a participant's hands were occluded by the head (e.g., when a participant leaned over the sink) as a result of the overhead perspective. Consideration of different mounting locations (e.g., offset from center may yield more favorable results system performance results.

Notwithstanding the limitations of this study, our results strongly support the furthered development of a depth-based hand tracker for the COACH prompting system. To this end, we now readdress our initial research questions:

1. *What effect will partially labeled training data have on the training and performance of decision trees?* Partially labeled data was shown to be an effective source of training data for a random decision forest. Similarly, the performance evaluation of decision forests was shown to be possible using the unweighted average recall metric.
2. *What training parameters impact the overall performance of the classifier?* We evaluated the impact of the maximum tree depth ($D_{max}$), the number of training images, the minimum cutoff gain ($g_{min}$), the maximum probe offset ($\theta_{max}$) and the number of training samples per image ($N$). Each of these parameters impacts the performance of the classifier. Optimal values were discovered for the maximum tree depth ($D_{max}$), the minimum cutoff gain ($g_{min}$), the maximum probe offset ($\theta_{max}$) and the number of training samples per image ($N$). When investigating the impact of the number of training images on classifier performance results suggested that an optimal value had not yet been, implying additional training images would improve performance.

3. *What is the mean average precision of classifier when proposing hand positions in three dimensions compared to ground-truth hand positions?* The final decision forest, trained using optimized training parameters, achieved a mean average precision of 0.846 over the *left hand, right hand,* and *head*, body parts

4. *How accurately can the COACH track the completion of task-based activities of users through the task of hand washing in a supervised real-world environment?* The COACH tracker classified the actions of the participants with a resulting $F_1$ measure score of 0.965 out of a maximum value of 1.

Future work will now look to address the limitations of the current study. A key component of this process will be to further test the efficacy of the new depth-based hand tracker through applications to different daily tasks, while simultaneously improving the classification ability by incorporating additional, diverse training data. Ultimately, the full integration of the new hand tracker into the COACH system will provide the foundation for additional, unsupervised, real-world efficacy studies of the COACH prompting system as an aid to older adults with dementia.

## V. ACKNOWLEDGEMENTS

The study was funded by grants from the Alzheimer Association (ETAC), Canadian Institute of Health Research (Operating Grant), and student funding from the NSERC CREATE CARE program.

9[25] P. Kohli and J. Shotton, "Key developments in human pose estimation for kinect," in *Consumer depth cameras for computer vision: Research topics and applications*, A. Fossati, J. Gall, H. Grabner, X. Ren, and K. Konolige, Eds., ed. London: Springer-Verlag, 2013.

[26] A. Kolb, E. Barth, R. Koch, and R. Larsen, "Time-of-flight cameras in computer graphics," *Computer Graphics Forum,* vol. 29, pp. 141-159, 2010.

[27] V. Megavannan, B. Agarwal, and R. Venkatesh Babu, "Human action recognition using depth maps," in *Signal Processing and Communications (SPCOM), 2012 International Conference on*, 2012, pp. 1-5.

[28] A. Kanaujia, N. Kittens, and N. Ramanathan, "Part segmentation of visual hull for 3D human pose estimation," presented at the Computer Vision and Pattern Recognition Workshops (CVPRW), 2013 IEEE Conference on, Portland, OR, 2013.

[29] I. Oikonomidis, N. Kyriazis, and A. A. Argyros, "Efficient model-based 3D tracking of hand articulations using Kinect," in *British Machine Vision Conference, BMVC 2011*, University of Dundee, UK, 2011.

[30] L. Cheng and K. M. Kitani, "Pixel-level hand detection for ego-centric videos," presented at the Conference on Computer Vision and Pattern Recognition (CVPR), 2013.

[31] X. Suau, J. Ruiz-Hidalgo, and J. R. Casas, "Real-Time Head and Hand Tracking Based on 2.5D Data," *Multimedia, IEEE Transactions on,* vol. 14, pp. 575-585, 2012.

[32] B. J. Southwell and G. Fang, "Human object recognition using colour and depth information from an RGB-D Kinect sensor," *International Journal of Advanced Robotic Systems,* vol. 10, 2013.

[33] OpenNI. (2012, December 11). *OpenNI Modules*. Available: http://openni.org/Downloads/OpenNIModules.aspx

[34] T. R. I. University Health Network. (2014, January 21). *UHN Toronto Rehabilitation Institute*. Available: http://www.uhn.ca/TorontoRehab

[35] Toronto Rehabilitation Institute. (2014, January 21). *HomeLab*. Available: http://www.idapt.com/index.php/labs-services/research-labs/homelab

[36] V. Labatut and H. Cherifi, "Evaluation of performance measures for classifiers comparions," *Ubiquitous Computing and Communication Journal,* vol. 6, pp. 21-31, 2011.

[37] B. Schuller, S. Steidl, and A. Batliner, "The INTERSPEECH 2009 Emotion Challenge," presented at the 10th Annual Conference of the International Speech Communication Association, Brighton, United Kingdom, 2009.

[38] K. Fukunaga and L. Hostetler, "The estimation of the gradient of a density function, with applications in pattern recognition," *IEEE Transactions on Information Theory,* vol. 21, pp. 32-40, 1975.